\documentclass[lettersize,journal]{IEEEtran}
\usepackage{amsmath,amsfonts}
\usepackage{algorithmic}
\usepackage{algorithm}
\usepackage{array}
\usepackage{multirow} 
\usepackage[caption=false,font=normalsize,labelfont=sf,textfont=sf]{subfig}
\usepackage{textcomp}
\usepackage{stfloats}
\usepackage{url}
\usepackage{verbatim}
\usepackage{graphicx}
\usepackage{cite}
\usepackage{makecell}
\usepackage{booktabs}
\begin{document}
\title{Cyclic Self-Supervised Diffusion for Ultra Low-field to High-field MRI Synthesis}
\author{Zhenxuan Zhang, Peiyuan Jing, Zi Wang, Ula Briski, Coraline Beitone, Yue Yang, Yinzhe Wu, Fanwen Wang, Liutao Yang, Jiahao Huang, Zhifan Gao, Zhaolin Chen, Kh Tohidul Islam, Guang Yang, Peter J. Lally
\thanks{Guang Yang was supported in part by the ERC IMI (101005122), the H2020 (952172), the MRC (MC/PC/21013), the Royal Society (IEC/NSFC/211235), the NVIDIA Academic Hardware Grant Program, the SABER project supported by Boehringer Ingelheim Ltd, NIHR Imperial Biomedical Research Centre (RDA01), The Wellcome Leap Dynamic resilience program (co-funded by Temasek Trust)., UKRI guarantee funding for Horizon Europe MSCA Postdoctoral Fellowships (EP/Z002206/1), UKRI MRC Research Grant, TFS Research Grants (MR/U506710/1), Swiss National Science Foundation (Grant No. 220785), and the UKRI Future Leaders Fellowship (MR/V023799/1). Zhenxuan Zhang was supported by a CSC Scholarship. Zhenxuan Zhang and Peiyuan Jing are co-first authors. Guang Yang and Peter J. Lally are co-last authors. Corresponding author: Guang Yang (\texttt{g.yang@imperial.ac.uk}).}%
\thanks{Zhenxuan Zhang, Peiyuan Jing, Zi Wang, Ula Briski, Yue Yang, Coraline Beitone, Fanwen Wang, Yinzhe Wu, Jiahao Huang, and Liutao Yang are with the Department of Bioengineering and Imperial-X, Imperial College London, London SW7 2AZ, U.K.}%
\thanks{Coraline Beitone is with the Department of Bioengineering, Imperial College London, London SW7 2AZ, U.K.}
\thanks{Peter J. Lally is with the Department of Bioengineering, Imperial College London, London SW7 2AZ, U.K. and the U.K. Dementia Research Institute Centre for Care Research \& Technology, London, W12 0BZ}
\thanks{Zhifan Gao is with the School of Biomedical Engineering, Sun
Yat-sen University, Shenzhen 518107, China}
\thanks{Zhaolin Chen and Kh Tohidul Islam are with the Monash Biomedical Imaging, Monash University, Clayton, VIC, Australia.}
\thanks{Guang Yang is with the Bioengineering Department and Imperial-X, Imperial College London, London W12 7SL, U.K., and with the National Heart and Lung Institute, Imperial College London, London SW7 2AZ, U.K., and with the Cardiovascular Research Centre, Royal Brompton Hospital, London SW3 6NP, U.K., and with the School of Biomedical Engineering \& Imaging Sciences, King’s College London, London WC2R 2LS, U.K.}}

\maketitle

\begin{abstract}
Synthesizing high-quality images from low-field MRI holds significant potential. Low-field MRI is cheaper, more accessible, and safer, but suffers from low resolution and poor signal-to-noise ratio. This synthesis process can reduce reliance on costly acquisitions and expand data availability. However, synthesizing high-field MRI still suffers from a clinical fidelity gap. There is a need to preserve anatomical fidelity, enhance fine-grained structural details, and bridge domain gaps in image contrast. To address these issues, we propose a \emph{cyclic self-supervised diffusion (CSS-Diff)} framework for high-field MRI synthesis from real low-field MRI data. Our core idea is to reformulate diffusion-based synthesis under a cycle-consistent constraint. It enforces anatomical preservation throughout the generative process rather than just relying on paired pixel-level supervision. The CSS-Diff framework further incorporates two novel processes. The slice-wise gap perception network aligns inter-slice inconsistencies via contrastive learning. The local structure correction network enhances local feature restoration through self-reconstruction of masked and perturbed patches. Extensive experiments on cross-field synthesis tasks demonstrate the effectiveness of our method, achieving state-of-the-art performance (e.g., 31.80 $\pm$ 2.70 dB in PSNR, 0.943 $\pm$ 0.102 in SSIM, and 0.0864 $\pm$ 0.0689 in LPIPS). Beyond pixel-wise fidelity, our method also preserves fine-grained anatomical structures compared with the original low-field MRI (e.g., left cerebral white matter error drops from 12.1$\%$ to 2.1$\%$, cortex from 4.2$\%$ to 3.7$\%$). To conclude, our CSS-Diff can synthesize images that are both quantitatively reliable and anatomically consistent.


\end{abstract}
\section{Introduction}
\begin{figure}[t]
\centerline{\includegraphics[width=\columnwidth]{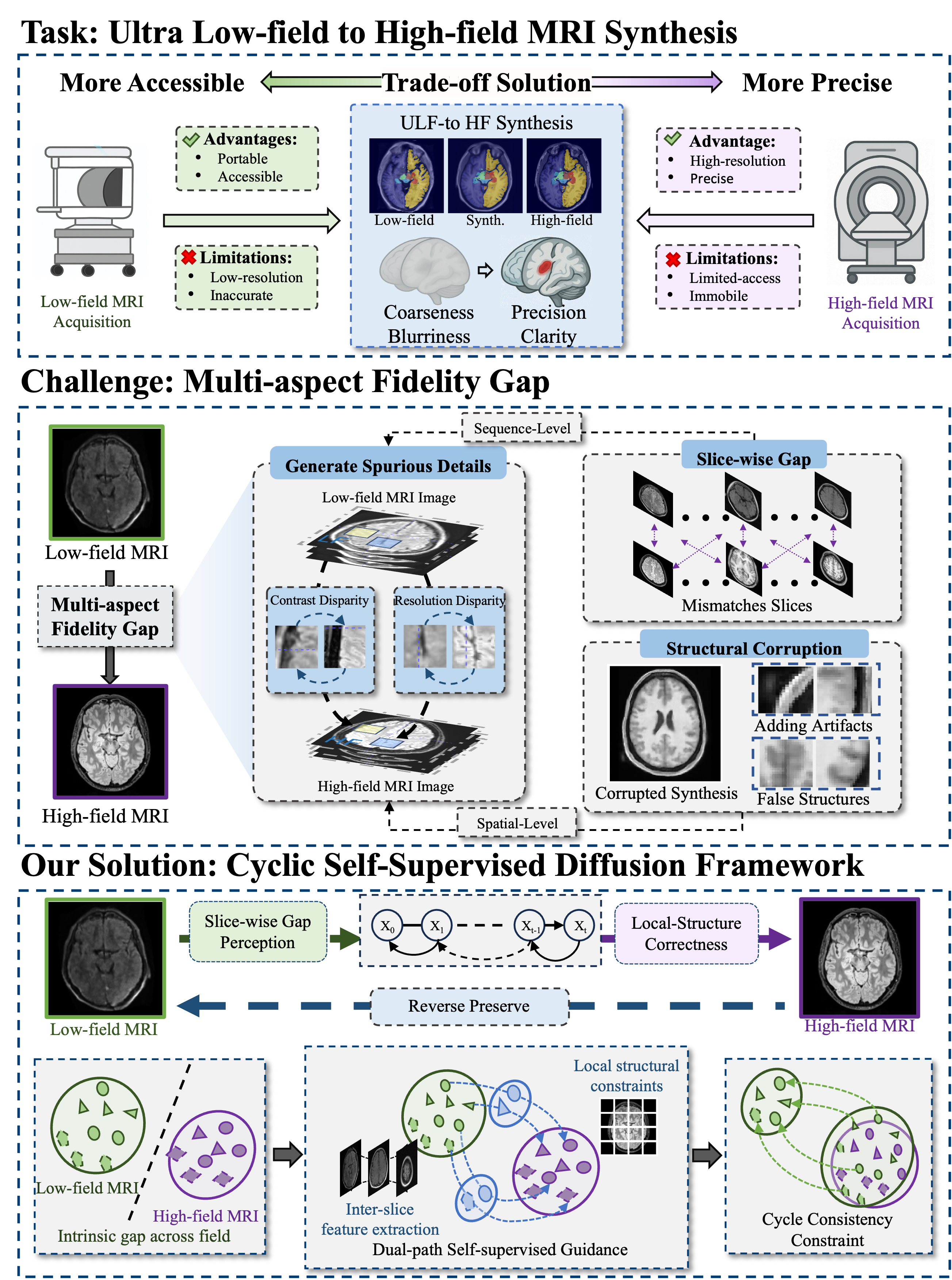}}
\caption{The motivation and challenges of the proposed CSS-Diff framework. (a) Motivation: Low-field MRI is portable but blurry and inaccurate, while high-field MRI is precise but costly and immobile. Synthesizing high-field quality from low-field inputs improves clarity and diagnostic reliability. (b) Multi-aspect fidelity gap: The task faces three challenges: spurious details from contrast–resolution disparity, slice-wise gaps from spatial mismatches, and structural corruption with artifacts or false patterns. (c) The CSS-Diff uses a reverse-preserve strategy with self-supervised guidance. It perceives slice-wise gaps, extracts inter-slice features, and enforces local structural constraints. This enables cycle-consistent synthesis of high-field MRI with preserved anatomical fidelity.}
\label{fig1}
\end{figure}

Synthesizing high-field-like MRI from low-field acquisitions offers a potential way to retain the accessibility of low-field imaging while improving its visual fidelity (Fig.~\ref{fig1} (a)). Magnetic resonance imaging (MRI) is essential for clinical diagnosis. High-field MRI gives high-quality images with detailed tissue contrast. But its use is limited by high purchase and maintenance costs, heavy infrastructure requirements, and poor suitability for deployment in remote or resource-limited settings. In contrast, low-field MRI (below 1.5 T) is cheaper and more portable. In many community hospitals, outpatient clinics, emergency departments and mobile imaging units, low-field MRI is often the only practical choice due to its lower cost, portability and minimal infrastructure requirements \cite{zhao2024whole,unity}. However, it has a low signal-to-noise ratio (SNR) and poor spatial resolution. Fine anatomical structures are hard to see. As a result, small lesions, subtle demyelination, and microvascular changes are often missed. These are important in diseases such as early-stage multiple sclerosis and small-vessel disease. Enhancing low-field MRI to produce high-field–like images can improve diagnostic accuracy. This also keeps the advantage of accessibility. Further, extending the synthesis to multiple field strengths and contrasts (T$_1$w, T$_2$w, FLAIR) can further support tasks such as tissue segmentation and quantitative analysis. Therefore, it is necessary to develop a synthesis algorithm that can enhance fidelity and improve the clinical utility of low-field scans.

However, synthesizing high-field MRI from low-field inputs remains technically challenging due to the multi-aspect fidelity gaps (Fig.~\ref{fig1} (b))~\cite{low-field,leiden_MRI,KCL_dataset}. It involves spurious detail generation, slice-wise mismatch, and anatomical structure corruption. First, spurious details occur because of the large contrast and resolution gap between field strengths. This leads to modeling instability (e.g., inconsistent texture and unstable boundaries). Especially, structures that are faint or invisible in low field (e.g., microvasculature or subtle edema) appear clearly in high field~\cite{lin2023low}. Without reliable cues, the model may hallucinate critical features. This results in implausible patterns or over-smoothed textures. Second, slice-wise mismatch can arise from spatial mismatches between corresponding anatomical positions~\cite{resvit,SACE}. Inconsistent positioning, patient motion, and specific distortions can cause the same slice index in low- and high-field scans to represent slightly different anatomies. Even sub-voxel misalignments may distort spatial correspondence, leading to artifacts or blending that compromise anatomical integrity. Third, preserving structural details is critical to diagnostic fidelity~\cite{UNest,Cyclegan}. Low-field MRI often suffers from low SNR and blurring, obscuring fine boundaries such as cortical layers, small lesions, or vessels~\cite{HCP}. High-field MRI reveals these features more clearly, but reconstructing them from degraded inputs is ill-posed. Errors can alter the shape, size, or texture of the lesion, risking false negatives or positives in diagnosis. Addressing these limitations requires precise slice-wise alignment, faithful preservation of structural details, and stable model training across domain shifts. Therefore, synthesis must balance anatomical detail with clinical realism while maintaining stable convergence.


Existing methods still struggle to address the multi-aspect fidelity gap in low-to-high-field MRI synthesis. Early pixel-level supervised regressors learn voxel-wise intensity mappings~\cite{de2022deep,resvit}. These mappings assume good anatomical correspondence. But contrast and resolution disparities obscure boundaries, which makes the correspondence imperfect~\cite{Cyclegan,cytran,UNest,yang2025unsupervised}. This may cause inconsistent voxel mappings and lead to misaligned slices. Therefore, these methods still cannot resolve the slice-wise alignment deficiency. GAN-based models optimize adversarial realism by matching the data distribution, yielding sharp textures and plausible appearance ~\cite{cytran,UNest}. These optimize visual realism but may sacrifice clinical fidelity~\cite{esrgan, SynGAN,yang2025unsupervised}. That is, images may look plausible but miss key diagnostic details and increase hallucination risk. Therefore, these GAN-based methods still cannot solve hallucination control and clinical fidelity. Cycle-consistent variants impose bidirectional constraints~\cite{Cyclegan,SynDiff}, which preserve content and mitigate gross misalignment. This encourages coarse anatomical consistency but micro-structures are not fully recovered. Therefore, these still cannot ensure fine-grained spatial–structural preservation. Diffusion-based models offer strong generative capacity and better fine-grained detail ~\cite{SynDiff,MiDiffusion,diffusion_survey,chung2022score}. But they may hallucinate plausible but false micro-structures without anatomical supervision. The hallucination control and slice-wise correspondence remain unresolved without alignment-aware anatomical constraints. Therefore, the key challenge remains to design a synthesis framework that explicitly accounts for slice-wise misalignment, spatial corruption and clinical fidelity.

\begin{figure*}[t]
\centerline{\includegraphics[width=2\columnwidth]{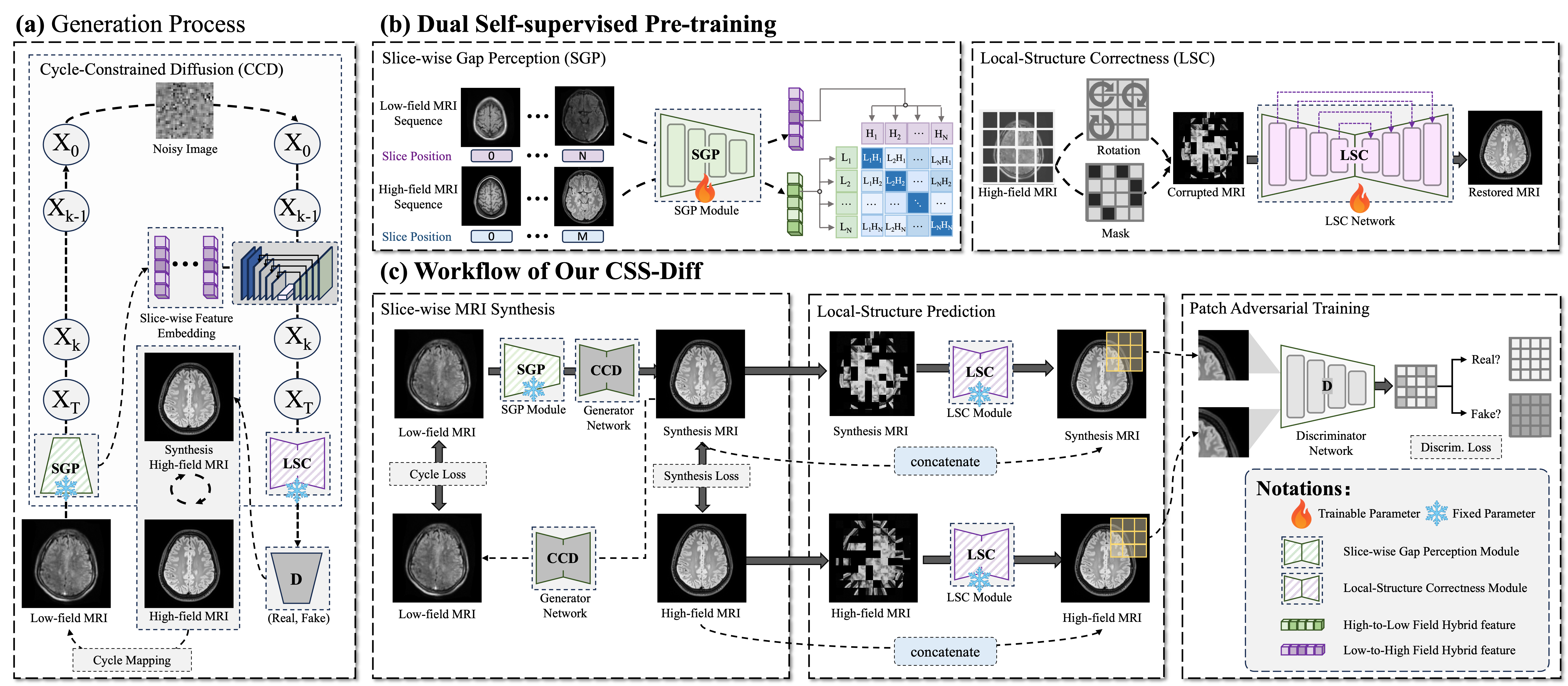}}
\caption{Detailed architecture of the proposed cyclic self-supervised diffusion (CSS-Diff) framework. (a) Progressive self-supervised diffusion gradually enhances MRI quality from low-field to high-field. (b) The framework incorporates slice-wise gap perception (SGP), local-structure correctness (LSC), and adversarial training to guide high-fidelity MRI synthesis. (c) Data Synthesis and Adversarial Training aims to synthesise high-field MRI data from low-field MRI inputs using a synthesis network.}
\label{fig2}
\end{figure*}
In this paper, we propose a Cyclic Self-Supervised Diffusion (CSS-Diff) framework for high-field MRI synthesis from low-field inputs (Fig.~\ref{fig1} (c)). Unlike prior approaches that rely on direct pixel-to-pixel supervision, CSS-Diff leverages diffusion trajectories as an iterative refinement process, where structural fidelity is progressively enhanced. To further regularize the transformation and ensure structural plausibility, CSS-Diff enforces a cycle-consistency constraint between the low-field input and the synthesized high-field output. In addition, inter-slice semantic consistency and local anatomical fidelity are jointly optimized during generation, enabling the model to recover volumetrically coherent and anatomically faithful structures. Our CSS-Diff explicitly addresses three major challenges in low-field to high-field MRI synthesis. (1) The CSS-Diff incorporates a cycle-consistency constraint within the diffusion trajectory to avoid synthesizing unrealistic structures. It enforces consistency between the low-field input and the synthesized high-field output. This regularizes the transformation and preserves structural plausibility. (2) The CSS-Diff introduces a slice-wise gap perception module to mitigate through-plane inconsistencies in conventional slice-wise synthesis. It leverages sequence-aware contrastive learning to capture dependencies between adjacent slices, which enhances inter-slice continuity and improves volumetric coherence during generation. (3) The CSS-Diff employs a local structure correction mechanism to reduce distortions in fine anatomical details during cross-field translation. It is based on self-supervised masked and rotated patch reconstruction, which exposes implausible textures and provides corrective feedback, enabling the model to faithfully recover subtle anatomical features. By jointly integrating cycle-constrained diffusion, slice-wise gap perception, and local structure correction, CSS-Diff achieves anatomically realistic and high-fidelity MRI synthesis in both internal and external datasets. The contribution lies in four folds:
\begin{enumerate}
    \item We design a diffusion-based framework that integrates cycle-consistency constraint, enabling bidirectional reconstruction between low-field and high-field domains. This ensures that the synthesized images remain faithful to the underlying anatomy.
    \item We propose a slice-wise gap perception mechanism that endows the model with position-specific awareness across the z-axis, enabling slice-dependent feature conditioning and better discrimination of anatomically adjacent slices.
    \item We propose a local structure correction strategy that selectively amplifies fine-grained structural errors during synthesis. This preserves subtle anatomical details and improves the clinical reliability of the generated MRIs.
    \item Our CSS-Diff is validated on three datasets with paired low-field to high-field MRI and proves effective across multiple contrasts (T$_1$w, T$_2$w, FLAIR). The enhanced images provide clearer delineation of critical structures such as the hippocampus, cortex, and thalamus.
\end{enumerate}

\section{Related Work}
\subsubsection{Image Synthesis Methods}

Early approaches to image synthesis in medical imaging focused on traditional interpolation, reconstruction, and model‐based methods \cite{modat2010fast,chen2017low,wang2025one}. These methods exploited known physics of the acquisition process. However, these techniques were often limited in their ability to recover fine structural details and textures from low-quality data. With the advent of deep learning, generative-based models have revolutionised image synthesis, such as variational autoencoders (VAEs) and generative adversarial networks (GANs) \cite{goodfellow2020generative,pix2pix,Cyclegan}. In particular, conditional and unpaired methods such as Pix2Pix and CycleGAN have been used extensively to perform cross-modal translations \cite{pix2pix,Cyclegan}. They augment limited clinical datasets while preserving anatomical details \cite{nie2017medical}. More recently, diffusion models have emerged as a promising alternative due to their ability to generate images with superior fidelity and diversity \cite{ddpm,diffusion_survey,chung2022score}. These models operate by gradually adding noise to images and then learning to reverse the process. These models provide a mechanism to synthesise high-quality images and show potential to meet clinical requirements. However, a major limitation of current generative approaches is the risk of hallucinating non-existent structures. This mainly comes from the lack of explicit anatomical or physical constraints. It raises concerns about their reliability in clinical use.

\subsubsection{Self-supervised Methods for Image Synthesis}
Self-supervised learning provides unique advantages for medical image synthesis, especially in scenarios where paired cross-modal data are limited. By leveraging proxy objectives such as contrastive learning, masked reconstruction, or context prediction, models can exploit abundant unlabeled data to learn modality-invariant yet anatomy-aware representations \cite{gui2024survey, zhou2019models}. This has two major benefits. First, it enhances structural fidelity by making the synthesized images more sensitive to subtle anatomical cues, thereby reducing hallucinations and mode collapse \cite{chaitanya2020contrastive}. Second, it facilitates domain transfer, as self-supervised features generalize better across field strengths and contrast. These properties make self-supervised synthesis particularly appealing for improving clinical reliability and downstream utility \cite{SACE}. Nonetheless, current self-supervised approaches also have limitations. Many pretext tasks are designed heuristically and may not align perfectly with diagnostic priorities; for example, patch-level contrastive objectives can improve texture realism but may neglect global tissue contrast. Moreover, while self-supervised constraints reduce artifacts, they can sometimes oversmooth fine structures or enforce excessive invariance, diminishing subtle pathological signals \cite{yang2020unsupervised}. These highlight that the pretext tasks must align with clinically relevant features and balance local fidelity with global realism.

\subsubsection{Cross-field MRI Analysis}
Low-field MRI is cheaper and more portable but suffers from noise, low resolution, and reduced diagnostic reliability. High-field MRI, by contrast, offers superior SNR and resolution, which are critical for subtle anatomical and pathological features \cite{sarracanie2020low,marques2019low,le2022magnetic}. This gap has motivated methods to synthesize high-field quality from low-field inputs. Early GAN-based approaches \cite{goodfellow2020generative,nie2017medical,isola2017image,yang2020unsupervised} improve perceptual realism but often introduce hallucinated details. Diffusion models \cite{ddpm,diffusion_survey} have recently been applied to MRI synthesis \cite{dayarathna2024ultra,SynDiff,MiDiffusion}, offering better stability and fidelity at higher computational cost. However, these existing studies rely on synthetic degradations \cite{leiden_MRI,KCL_dataset, al2023usability} that may not capture true low-field features. However, a challenge remains the balance between fidelity and generative performance (i.e., synthesized images must preserve anatomical accuracy while also enhancing resolution and visual quality for reliable downstream use)\cite{lin2023low,dayarathna2024ultra}. Achieving this balance is difficult because fine details may be lost in low-field inputs, making it unclear whether generated structures reflect true anatomy or hallucinated content \cite{nie2017medical,isola2017image}. Moreover, clinically relevant features such as small lesions are especially vulnerable to distortion during synthesis, which further complicates reliable translation \cite{ddpm,SynDiff,MiDiffusion}.
\begin{table*}[t!]
\centering
\caption{Evaluation on the paired 64mT$\rightarrow$3T dataset under Multi-Contrast (T$_1$w, T$_2$w, FLAIR) setting. Bold indicates our CSS-Diff results. Results are reported on both internal (partially seen in training) and external (unseen in training) datasets. $^{*}$ indicates $p < 0.05$ and $^{**}$ indicates $p < 0.01$ in a Wilcoxon signed-rank test against our CSS-Diff model.}
\label{tab:comparison}
\resizebox{\textwidth}{!}{
\begin{tabular}{l|l|c|ccc|ccc}
\hline\hline
\multirow{2}{*}{\textbf{Setting}} & \multirow{2}{*}{\textbf{Method}} & \multirow{2}{*}{\textbf{Year}} & \multicolumn{3}{c|}{\textbf{Monash Uni. Dataset (Internal)}} & \multicolumn{3}{c}{\textbf{Leiden Uni. Dataset (Internal)}} \\
\cline{4-9}
 & & & PSNR$\uparrow$ & SSIM$\uparrow$ & LPIPS$\downarrow$ & PSNR$\uparrow$ & SSIM$\uparrow$ & LPIPS$\downarrow$ \\
\hline
&Low-field & - & 21.12$\pm$2.27$^{**}$ & 0.731$\pm$0.147$^{**}$ & 0.2449$\pm$0.0874$^{**}$ & 21.16$\pm$2.34$^{**}$ & 0.736$\pm$0.139$^{**}$ & 0.2434$\pm$0.0848$^{**}$ \\
\hline
\multirow{3}{*}{\textbf{Unpaired}} 
 & CycleGAN\cite{Cyclegan}  & 2017 & 24.74$\pm$2.17$^{**}$ & 0.794$\pm$0.154$^{**}$ & 0.1891$\pm$0.0849$^{**}$ & 24.86$\pm$2.27$^{**}$ & 0.800$\pm$0.141$^{**}$ & 0.1888$\pm$0.0843$^{**}$ \\
 & SynGAN\cite{SynGAN}        & 2021 & 24.40$\pm$6.55$^{**}$ & 0.709$\pm$0.289$^{**}$ & 0.3359$\pm$0.1802$^{**}$ & 24.14$\pm$6.56$^{**}$ & 0.697$\pm$0.295$^{**}$ & 0.3421$\pm$0.1755$^{**}$ \\
 & UNest\cite{UNest}          & 2024 & 23.10$\pm$2.42$^{**}$ & 0.763$\pm$0.163$^{**}$ & 0.2327$\pm$0.0852$^{**}$ & 23.15$\pm$2.51$^{**}$ & 0.770$\pm$0.151$^{**}$ & 0.2301$\pm$0.0857$^{**}$ \\
\hline
\multirow{8}{*}{\textbf{Paired}} 
 & Pix2Pix\cite{isola2017image} & 2017 & 29.24$\pm$2.43$^{**}$ & 0.918$\pm$0.109$^{**}$ & 0.1100$\pm$0.0512$^{**}$ & 29.44$\pm$2.69$^{**}$ & 0.924$\pm$0.088$^{**}$ & 0.1075$\pm$0.0435$^{**}$ \\
 & ESRGAN\cite{esrgan}        & 2018 & 29.59$\pm$3.36$^{**}$ & 0.920$\pm$0.119$^{**}$ & 0.1125$\pm$0.0873$^{**}$ & 29.75$\pm$3.55$^{**}$ & 0.926$\pm$0.097$^{**}$ & 0.1090$\pm$0.0866$^{**}$ \\
 & TranUnet\cite{transunet}   & 2021 & 30.42$\pm$2.81$^{*}$  & 0.927$\pm$0.119$^{*}$  & 0.0973$\pm$0.0720$^{*}$  & 30.67$\pm$3.05$^{*}$  & 0.933$\pm$0.111$^{*}$  & 0.0955$\pm$0.0730$^{*}$  \\
 & ResViT\cite{resvit}        & 2022 & 30.45$\pm$2.69$^{**}$ & 0.930$\pm$0.106$^{**}$ & 0.0921$\pm$0.0660$^{**}$ & 30.71$\pm$2.89$^{**}$ & 0.937$\pm$0.084$^{**}$ & 0.0891$\pm$0.0627$^{**}$ \\
 & CyTran\cite{cytran}        & 2023 & 31.21$\pm$3.05$^{*}$  & 0.940$\pm$0.100$^{*}$  & 0.0974$\pm$0.0550$^{*}$  & 31.41$\pm$3.40$^{*}$  & 0.946$\pm$0.079$^{*}$  & 0.0965$\pm$0.0544$^{*}$ \\
 & SynDiff\cite{SynDiff}      & 2023 &30.44$\pm$2.48$^{**}$ & 0.929$\pm$0.120$^{**}$ & 0.1190$\pm$0.0418$^{**}$ & 30.39$\pm$2.36$^{**}$ & 0.928$\pm$0.110$^{**}$ & 0.1214$\pm$0.0425$^{**}$ \\
 & MiDiffusion\cite{MiDiffusion} & 2024 & 30.09$\pm$4.37$^{**}$ & 0.912$\pm$0.105$^{**}$ & 0.1255$\pm$0.0676$^{**}$ & 30.25$\pm$4.35$^{**}$ & 0.917$\pm$0.089$^{**}$ & 0.1218$\pm$0.0609$^{**}$ \\
 & \textbf{CSS-Diff}                   &      & \textbf{31.80$\pm$2.70} & \textbf{0.943$\pm$0.102} & \textbf{0.0864$\pm$0.0689} & \textbf{31.96$\pm$2.88} & \textbf{0.948$\pm$0.083} & \textbf{0.0850$\pm$0.0624} \\
\hline
\multirow{2}{*}{\textbf{Setting}} & \multirow{2}{*}{\textbf{Method}} & \multirow{2}{*}{\textbf{Year}} & \multicolumn{3}{c|}{\textbf{KCL Uni. ses-HFE Dataset (External)}} & \multicolumn{3}{c}{\textbf{KCL Uni. ses-HFC Dataset (External)}} \\
\cline{4-9}
 & & & PSNR$\uparrow$ & SSIM$\uparrow$ & LPIPS$\downarrow$ & PSNR$\uparrow$ & SSIM$\uparrow$ & LPIPS$\downarrow$ \\
\hline
&Low-field & - & 23.32$\pm$1.85$^{**}$ & 0.757$\pm$0.108$^{**}$ & 0.2042$\pm$0.0509$^{**}$ & 22.90$\pm$1.98$^{**}$ & 0.729$\pm$0.122$^{**}$ & 0.2182$\pm$0.0605$^{**}$ \\
\hline
\multirow{3}{*}{\textbf{Unpaired}} 
 & CycleGAN\cite{Cyclegan}    & 2017 & 26.27$\pm$1.54$^{**}$ & 0.756$\pm$0.143$^{**}$ & 0.1950$\pm$0.0573$^{*}$  & 25.64$\pm$1.62$^{**}$ & 0.748$\pm$0.179$^{**}$ & 0.2007$\pm$0.0621$^{*}$ \\
 & SynGAN\cite{SynGAN}        & 2021 & 26.54$\pm$1.57$^{**}$  & 0.781$\pm$0.160$^{**}$  & 0.2496$\pm$0.0897$^{**}$ & 26.16$\pm$1.76$^{**}$  & 0.756$\pm$0.181$^{**}$  & 0.2614$\pm$0.0967$^{**}$ \\
 & UNest\cite{UNest}          & 2024 & 26.12$\pm$1.53$^{**}$ & 0.774$\pm$0.177$^{**}$  & 0.2104$\pm$0.0507$^{**}$  & 25.73$\pm$1.70$^{**}$ & 0.750$\pm$0.191$^{**}$  & 0.2260$\pm$0.0642$^{**}$ \\
\hline
\multirow{8}{*}{\textbf{Paired}} 
 & Pix2Pix\cite{isola2017image} & 2017 & 26.64$\pm$1.72$^{**}$ & 0.770$\pm$0.165$^{**}$ & 0.2015$\pm$0.0513$^{*}$  & 26.19$\pm$1.84$^{**}$ & 0.743$\pm$0.178$^{**}$ & 0.2150$\pm$0.0629$^{**}$ \\
 & ESRGAN\cite{esrgan}        & 2018 & 26.08$\pm$2.10$^{**}$ & 0.780$\pm$0.168$^{**}$ & 0.2132$\pm$0.0781$^{**}$ & 25.59$\pm$2.15$^{**}$ & 0.749$\pm$0.185$^{**}$ & 0.2292$\pm$0.0823$^{**}$ \\
 & TranUnet\cite{transunet}   & 2021 & 26.85$\pm$1.50$^{**}$  & 0.799$\pm$0.165$^{*}$  & 0.1859$\pm$0.0501        & 25.94$\pm$1.57$^{*}$  & 0.730$\pm$0.163$^{**}$ & 0.2093$\pm$0.0662$^{*}$ \\
 & ResViT\cite{resvit}        & 2022 & 26.77$\pm$1.58$^{**}$  & 0.791$\pm$0.161$^{*}$  & 0.1921$\pm$0.0528$^{*}$        & 26.29$\pm$1.73$^{**}$  & 0.762$\pm$0.174$^{*}$  & 0.2076$\pm$0.0651$^{*}$ \\
 & CyTran\cite{cytran}        & 2023 & 25.95$\pm$1.47$^{**}$ & 0.772$\pm$0.162$^{**}$ & 0.1886$\pm$0.0536  & 25.64$\pm$1.62$^{**}$ & 0.748$\pm$0.179$^{**}$ & 0.2007$\pm$0.0621$^{**}$ \\
 & SynDiff\cite{SynDiff}      & 2023 & 23.94$\pm$1.86$^{**}$ & 0.793$\pm$0.144$^{**}$ & 0.2061$\pm$0.0519$^{**}$ & 23.47$\pm$1.72$^{**}$ & 0.764$\pm$0.147$^{**}$ & 0.2230$\pm$0.0639$^{**}$ \\
 & MiDiffusion\cite{MiDiffusion} & 2024 & 26.11$\pm$1.67$^{**}$ & 0.779$\pm$0.162$^{**}$ & 0.1973$\pm$0.0502$^{*}$  & 25.70$\pm$1.81$^{**}$ & 0.754$\pm$0.174$^{**}$ & 0.2101$\pm$0.0594$^{**}$\\
 & \textbf{CSS-Diff} &      & \textbf{27.25$\pm$1.61} & \textbf{0.807$\pm$0.129} & \textbf{0.1901$\pm$0.0626} & \textbf{26.97$\pm$1.70} & \textbf{0.785$\pm$0.144} & \textbf{0.2006$\pm$0.0676} \\
\hline\hline
\end{tabular}}
\end{table*}

\section{Method}
\subsection{Problem Formulation}

Magnetic field strength (\( B_0 \)) fundamentally determines MRI signal characteristics. High-field MRI (\( B_0 \geq 3\,\mathrm{T} \)) offers a higher SNR, resolution, and contrast, while low-field MRI (\( B_0 \leq 0.5\,\mathrm{T} \)) is more accessible but suffers from degraded anatomical fidelity, especially in fine structures such as lesions or vessel boundaries.

Empirical analysis has shown that, under typical acquisition and hardware conditions, SNR increases approximately quadratically with magnetic field strength (\( \mathrm{SNR} \propto B_0^2 \)) \cite{le2022magnetic}, and is also proportional to the acquired voxel volume. Low-field MRI, therefore, has much lower inherent SNR, which is often partially compensated by increasing voxel size at the expense of spatial resolution. This trade-off limits the ability to resolve small structures that may be essential for clinical interpretation. To address this, we aim to synthesize high-field-like images from low-field inputs, enhancing anatomical clarity and diagnostic utility (Fig.~\ref{fig2}(a)).

Let \(X \sim p_X(x \mid B_L)\) and \(Y \sim p_Y(y \mid B_H)\) represent magnitude image distributions at low and high field strengths, respectively. We seek a mapping \(G_\theta: X \rightarrow Y'\) such that \(Y' \approx Y\) in both semantic structure and fidelity, where \(\theta\) denotes parameters of the synthesis model. This can be formulated as
\begin{equation}
\theta^{*} = \arg\min_{\theta \in \Theta} \; \mathcal{L}(G_\theta),
\quad \mathcal{H} = \{ G_\theta: X \rightarrow Y' \},
\end{equation}
where the synthesis task $\mathcal{L}(G_\theta)$ integrates hallucination suppression, slice-wise alignment, and spatial detail preservation with appropriate weighting.

(i) Hallucination in generation process: Generative models may introduce spurious structures not supported by the low-field anatomy. We decompose the output into an anatomy-traceable part $A_\theta(x)$ and a hallucination residual $h_\theta(x)$,
\begin{equation}
G_\theta(x) \;=\; A_\theta(x) \;+\; h_\theta(x),
\end{equation}
where \(A_\theta(x)\) denotes structures that are consistent with the input anatomy and \(h_\theta(x)\) collects unsupported details. Introducing a reverse mapping \(F\) back to the low-field domain yields the cycle error.
\begin{equation}
E_{\text{cyc}}(x) \;=\; \|F(G_\theta(x)) - x\|_1.
\end{equation}
If mapping \(F\) is locally bi-Lipschitz with constant \(m>0\), then for \(u=A_\theta(x)\), \(v=A_\theta(x)+h_\theta(x)\),
\begin{equation}
\|F(G_\theta(x)) - x\|_1 \;=\; \|F(v)-F(u)\|_1 \;\ge\; m\,\|h_\theta(x)\|_1,
\end{equation}
where \(m>0\) is the local bi-Lipschitz lower constant of \(F\) near \(u,v\). So minimizing \(E_{\text{cyc}}\) effectively suppresses the hallucination residual \(\|h_\theta(x)\|_1\), guiding the generator toward anatomy-traceable solutions.

(ii) Slice-wise misalignment.
Low- and high-field scans may differ by discrete through-plane shifts or re-indexing along the $z$-axis. Without explicit slice information, $G_\theta$ cannot disambiguate adjacent slices. We model the unknown slice mapping by a re-indexing operator $S \sim \mathcal{S}$ and define the expected alignment error.
\begin{equation}
E_{\text{align}}(x,y)
=
\mathbb{E}_{S \sim \mathcal{S}}
 \,\big\| G_\theta(x) - S[y] \big\|_2^2
\end{equation}
Minimizing $E_{\text{align}}$ is ill-posed without slice cues. Adjacent slices are highly correlated, so multiple re-indexings $S$ can yield similar loss. Without an explicit slice identity signal, the mapping is non-identifiable.

(iii) Spatial-structure degradation.
During synthesis, existing anatomy may be altered. We describe this as a structural corruption in the generated image:
\begin{equation}
G_\theta(x) \;=\; y \circ \phi_\theta \;+\; \eta_\theta,
\end{equation}
where $\phi_\theta$ is an in-plane deformation and $\eta_\theta$ is an appearance residual. Structural degradation occurs when $\phi_\theta$ deviates from the identity or when $\eta_\theta$ removes or invents fine structures. This captures changes to anatomical boundaries and fine spatial detail arising during generation.

\subsection{Cycle-Constrained Diffusion}
To preserve anatomical structures in low-field MRI and reduce hallucinated details, we propose a Cycle-Constrained Diffusion (CCD) module (Fig.~\ref{fig2}(a)). It preserves anatomical structures in low-field MRI by enforcing cycle consistency between domains and path consistency along the diffusion trajectory, reducing hallucinations and stabilizing synthesis.

We model the low-to-high field MRI translation as a chain of $T$ generators:
\begin{equation}
    x_{t+1} = G_t(x_t, z_t; \theta_t), \quad z_t \sim \mathcal{N}(0, I), \quad t=0,\dots,T-1,
\end{equation}
yielding the final output: $ x_T = G_{T-1}\circ \cdots \circ G_0(x_0, z),$ where $\circ$ denotes function composition and $x_t$ denotes the image at step $t$. $z_t$ is a stochastic latent variable, and $\theta_t$ are the parameters of the $t$-th generator. 

A discriminator $D$ distinguishes synthesized $x_T$ from real high-field data $y \sim p_Y(y \mid B_H)$:
\begin{align}
    \mathcal{L}_{adv}^D &= - \mathbb{E}_{y}[\log D(y)] - \mathbb{E}_{x_0, z}[\log (1 - D(x_T))], \\
    \mathcal{L}_{adv}^G &= - \mathbb{E}_{x_0, z}[\log D(x_T)].
\end{align}

To suppress hallucinated structures not supported by the input, we introduce a reverse generator $F(\cdot;\phi)$ mapping synthesized images back to the source domain:
\begin{equation}
    \tilde{x}_0 = F(x_T;\phi).
\end{equation}
A cycle loss ensures reversibility:
\begin{equation}
\begin{aligned}
    \mathcal{L}_{cyc} = &\mathbb{E}_{x_0}[\|x_0 - F(G(x_0))\|_1] \\&+ \rho \, \mathbb{E}_{y}[\|y - G(F(y))\|_1].
\end{aligned}
\end{equation}
This penalizes structures that cannot be consistently mapped back, effectively reducing hallucinations.

\begin{figure*}[t]
\centerline{\includegraphics[width=2.1\columnwidth]{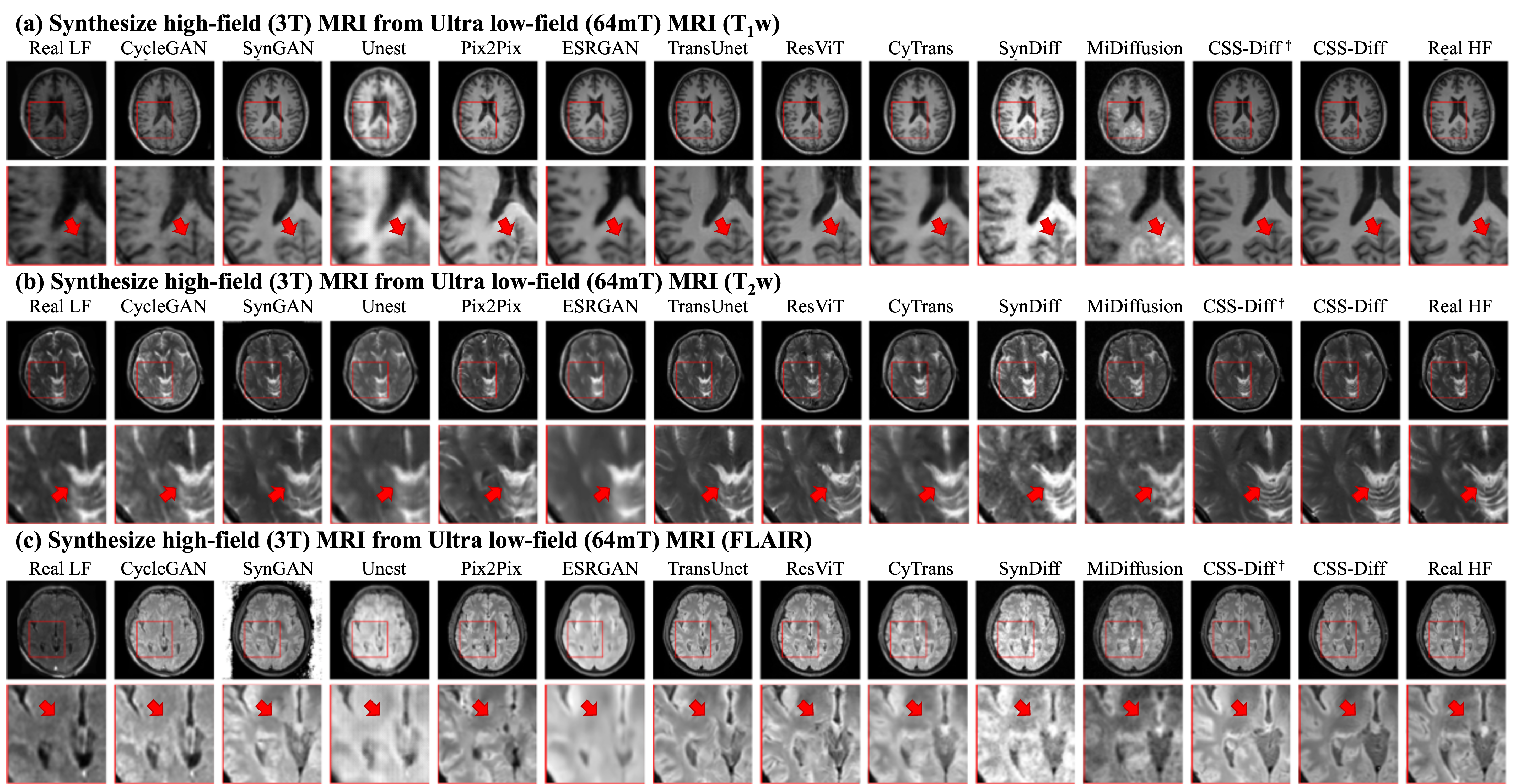}}
\caption{Visualization result of different baselines for exemplar regions. (top and bottom row of each panel, CSS-Diff$^\dagger$ denotes the CSS-Diff baseline model, while CSS-Diff indicates CSS-Diff with all modules enabled) (a) Synthesis of high-field MRI data from cross-contrast low-field MRI data. (b) Synthesis of ultra high-field MRI from same-contrast low-field MRI.}
\label{fig4}
\end{figure*}

To further constrain the internal trajectory, we adopt a diffusion forward process:
\begin{align}
q(x_t \mid x_{t-1}) &= \mathcal{N}\!\big(\sqrt{\alpha_t}\,x_{t-1},\, (1-\alpha_t)I\big),\\
q(x_t \mid x_0) &= \mathcal{N}\!\big(\sqrt{\bar{\alpha}_t}\,x_0,\, (1-\bar{\alpha}_t)I\big),
\end{align}
where $\alpha_t = 1-\beta_t$ and $\bar{\alpha}_t=\prod_{s=1}^t \alpha_s$. Given a noise predictor $\epsilon_\theta(x_t,t)$, one can estimate
\begin{equation}
    \hat{x}_0(x_t,t) \;=\; \frac{x_t - \sqrt{1-\bar{\alpha}_t}\,\epsilon_\theta(x_t,t)}{\sqrt{\bar{\alpha}_t}}.
\end{equation}
A deterministic update from $t$ to $t-1$ with $\eta=0$ can then be written as
\begin{equation}
    \tilde{x}_{t-1} \;=\; \sqrt{\bar{\alpha}_{t-1}}\,\hat{x}_0(x_t,t)
    \;+\; \sqrt{1-\bar{\alpha}_{t-1}}\;\epsilon_\theta(x_t,t).
\end{equation}

We define a path consistency loss by aligning the chain state $x_{t-1}$ with this deterministic reference $\tilde{x}_{t-1}$:
\begin{equation}
    \mathcal{L}_{path} = \sum_{t=1}^{T-1} \eta_t \, \mathbb{E}_{x_0, z}[\|x_{t-1} - \tilde{x}_{t-1}\|_2^2].
\end{equation}
This encourages the learned trajectory to remain reversible and stable, preventing divergence and suppressing unrealistic hallucinations.

The final optimization problem jointly updates the forward chain $\{G_t\}$ and reverse mapping $F$, while training the discriminator $D$ adversarially:
\begin{align}
\max_{\psi} \min_{\{\theta_t\},\,\phi}\ \mathcal{L}_{adv}^D+
\lambda_{1}\,\mathcal{L}_{adv}^G
+\lambda_{2}\,\mathcal{L}_{cyc}
+\lambda_{3}\,\mathcal{L}_{path},
\end{align}
where $\lambda_{1},\lambda_{2},\lambda_{3}\ge 0$ are loss-balancing coefficients. The cycle loss enforces reversibility and constrains hallucinations, while the DDIM-based path consistency term preserves diffusion-style progression. This balances realism, anatomical consistency, and trajectory stability.

\subsection{Dual Self-supervised Pretrain}
To further enhance realism, anatomical consistency, and trajectory stability, a self-supervised loop is incorporated into CSS-Diff. This loop introduces two dedicated modules: Slice-wise Gap Perception to mitigate inter-slice inconsistencies, and Local Structure Correction to refine fine-grained anatomical details (Fig.~\ref{fig2}(b)).
\subsubsection{Slice-wise Gap Perception}
Low-field and high-field MRI acquisitions inevitably exhibit slice-wise inconsistencies. Such inconsistencies break the through-plane anatomical continuity and hinder reliable synthesis across field strengths. To address this, we introduce a Slice-wise Gap Perception (SGP) module that provides a slice-level anatomical alignment constraint within the diffusion process.

The SGP leverages feature similarity to identify the most plausible high-field counterpart. Given a feature encoder $f(\cdot)$, the positive pair is defined as:
\begin{equation}
i^* = \arg\max_{j \in \{1,\dots,N\}} \sigma\!\left(f(S_{\mathrm{LF}}^{(i)}), f(S_{\mathrm{HF}}^{(j)})\right),
\end{equation}
where $\sigma(\cdot,\cdot)$ denotes cosine similarity. The resulting pair $\bigl(S_{\mathrm{LF}}^{(i)}, S_{\mathrm{HF}}^{(i^*)}\bigr)$ is treated as anatomically consistent, while all other $S_{\mathrm{HF}}^{(j)}$ $(j \neq i^*)$ act as negatives. 

A contrastive loss is then imposed to encourage slice-level consistency:
\begin{equation}
\mathcal{L}_{\text{SGP}} = -\frac{1}{N} \sum_{i=1}^N 
\log \frac{\exp\!\left(\sigma\!\left(f(S_{\mathrm{LF}}^{(i)}), f(S_{\mathrm{HF}}^{(i^*)})\right)/\tau \right)}
{\sum_{j=1}^N \exp\!\left(\sigma\!\left(f(S_{\mathrm{LF}}^{(i)}), f(S_{\mathrm{HF}}^{(j)})\right)/\tau \right)},
\end{equation}
where $\tau$ is a temperature scaling factor.

This objective produces slice-guided embeddings $z_i=f(S_{\mathrm{LF}}^{(i)})$ that encode slice identity, relative order, and local through-plane context across field strengths.
We embed $z_i$ into the generative model as conditioning for diffusion-based synthesis,
\begin{equation}
\epsilon_\theta(x_t, t \mid z_i),
\end{equation}
where $t\in\{1,\dots,T\}$ denotes the diffusion step in the reverse process, and $x_t$ is the intermediate state at step $t$. Thus, the contrastively learned features act as a soft slice prior inside the denoiser, guiding the trajectory toward anatomically coherent and through-plane consistent high-field MRI.

\subsubsection{Local Structure Correction}
To endow the model with the ability to perceive fine-grained anatomical structures, we design a self-supervised pretext task in the spirit of masked autoencoding. Given a high-field MRI data $Y$, we divide it into non-overlapping local blocks and deliberately perturb their spatial coherence through two transformations: (i) \emph{random rotation} $\psi_{\mathrm{rot}}$, which rotates a subset of blocks by $90^\circ$, $180^\circ$, or $270^\circ$; and (ii) \emph{random masking} $\phi_{\mathrm{mask}}$, which occludes another subset of blocks. The corrupted input is denoted as
\begin{equation}
y_{\text{LSC}} = \phi_{\mathrm{mask}}\!\left(\psi_{\mathrm{rot}}(y_{\text{syn}})\right).
\end{equation}

A reconstruction network $E_T(\cdot)$ is trained to restore the original image from $y_{\text{LSC}}$, thereby forcing the encoder to capture local anatomical priors. The reconstruction objective is defined as:
\begin{equation}
\mathcal{L}_{\text{LSC}} 
= \| y - E_T(y_{\text{LSC}}) \|_2
+ \alpha \big(1 - \mathrm{SSIM}(y, E_T(y_{\text{LSC}}) \big),
\end{equation}
where $E_T(y_{\text{LSC}})$ is the corrected image and $\alpha$ is a loss-balancing term. 

Futher, we use a patch discriminator with a compact adversarial objective:
\begin{equation}
\begin{aligned}
\min_{E_T}\ \max_{D}\ 
& \mathbb{E}_{y,(i,j)}\!\big[\log D(y)_{i,j}\big]
\\&+\mathbb{E}_{y_{\text{LSC}},(i,j)}\!\big[\log\!\big(1 - D\big(E_T(y_{\text{LSC}})\big)_{i,j}\big)\big],
\end{aligned}
\end{equation}
where $(i,j)$ indexes local patches and $D(\cdot)_{i,j}\in[0,1]$ is the realism score of the $(i,j)$-th patch.
\begin{table}[t]
\centering
\caption{Module ablation of the CSS-Diff framework ($\checkmark$ with module; $\times$ without module during training).}
\label{tab:module_ablation}
\resizebox{\columnwidth}{!}{
\begin{tabular}{c|ccc|ccc}
\hline\hline
\hline
\textbf{Setting} & \textbf{SGP} & \textbf{LSC} & \textbf{CCD} &
\textbf{PSNR$\uparrow$} & \textbf{SSIM$\uparrow$} & \textbf{LPIPS$\downarrow$} \\\hline
\multirow{4}{*}{\makecell[c]{Monash Uni.\\ dataset \\ (Internal)}}
 & $\times$ & $\times$ & $\times$ & 29.96$\pm$2.57 & 0.874$\pm$0.130 & 0.1099$\pm$0.0799 \\
 & $\checkmark$ & $\times$ & $\times$ & 30.07$\pm$2.57 & 0.895$\pm$0.138 & 0.1195$\pm$0.0827 \\
 & $\times$ & $\checkmark$ & $\times$ & 30.52$\pm$2.47 & 0.929$\pm$0.124 & 0.1123$\pm$0.0871 \\
 & $\checkmark$ & $\checkmark$ & $\checkmark$ & \textbf{32.44}$\pm$\textbf{2.74} & \textbf{0.940}$\pm$\textbf{0.109} & \textbf{0.0857}$\pm$\textbf{0.0711} \\
\hline
&           &           & &
\textbf{PSNR$\uparrow$} & \textbf{SSIM$\uparrow$} & \textbf{LPIPS$\downarrow$} \\\hline
\multirow{4}{*}{\makecell[c]{Leiden Uni.\\ dataset \\ (Internal)}}
 & $\times$ & $\times$ & $\times$ & 29.96$\pm$2.63 & 0.875$\pm$0.128 & 0.1074$\pm$0.0677 \\
 & $\checkmark$ & $\times$ & $\times$ & 30.10$\pm$2.60 & 0.898$\pm$0.130 & 0.1155$\pm$0.0697 \\
 & $\times$ & $\checkmark$ & $\times$ & 30.55$\pm$2.52 & 0.926$\pm$0.117 & 0.1082$\pm$0.0739 \\
 & $\checkmark$ & $\checkmark$ & $\checkmark$ & \textbf{32.38}$\pm$\textbf{2.77} & \textbf{0.937}$\pm$\textbf{0.117} & \textbf{0.0829}$\pm$\textbf{0.0548} \\
\hline
&           &           & & \textbf{PSNR$\uparrow$} & \textbf{SSIM$\uparrow$} & \textbf{LPIPS$\downarrow$} \\\hline
\multirow{4}{*}{\makecell[c]{KCL Uni.\\ dataset \\ (External)}}
 & $\times$ & $\times$ & $\times$ & 25.26$\pm$1.69 & 0.661$\pm$0.167 & 0.2152$\pm$0.0688 \\
 & $\checkmark$ & $\times$ & $\times$ & 26.88$\pm$1.50 & 0.745$\pm$0.175 & 0.2133$\pm$0.0704 \\
 & $\times$ & $\checkmark$ & $\times$ & 27.02$\pm$1.69 & 0.764$\pm$0.169 & 0.2204$\pm$0.0945 \\
 & $\checkmark$ & $\checkmark$ & $\checkmark$ & \textbf{27.02}$\pm$\textbf{1.66} & \textbf{0.762}$\pm$\textbf{0.183} & \textbf{0.1954}$\pm$\textbf{0.0666} \\
\hline\hline
\end{tabular}
}
\end{table}

\begin{table}[t]
\centering
\scriptsize
\caption{Dice scores of anatomical regions comparing low-field MRI and synthesised images against high-field ground truth. Values are reported as mean$_{\text{std}}$. Left and right hemispheric regions are merged. WM: white matter; GM: grey matter; CSF: cerebrospinal fluid; DC: diencephalon.}
\resizebox{\columnwidth}{!}{
\begin{tabular}{l|cc | l|cc}
\hline\hline
\textbf{Region} &  \textbf{LF (64mT)} & \textbf{Synthesised} & \textbf{Region} &  \textbf{LF (64mT)} & \textbf{Synthesised} \\
\hline\hline
WM &  0.75$_{\pm\text{0.03}}$ & 0.82$_{\pm\text{0.02}}$ 
& Inf. Lat. Ventricle &  0.24$_{\pm\text{0.12}}$ & 0.47$_{\pm\text{0.13}}$ \\
Cortical GM  &  0.65$_{\pm\text{0.03}}$ & 0.75$_{\pm\text{0.03}}$
& Cerebellum WM &  0.73$_{\pm\text{0.06}}$ & 0.77$_{\pm\text{0.05}}$ \\
CSF &  0.53$_{\pm\text{0.04}}$ & 0.62$_{\pm\text{0.04}}$
& Cerebellum GM &  0.79$_{\pm\text{0.07}}$ & 0.84$_{\pm\text{0.04}}$ \\
Hippocampus  &  0.69$_{\pm\text{0.08}}$ & 0.79$_{\pm\text{0.08}}$
& Pallidum  &  0.46$_{\pm\text{0.10}}$ & 0.71$_{\pm\text{0.08}}$ \\
Amygdala  &  0.71$_{\pm\text{0.09}}$ & 0.81$_{\pm\text{0.08}}$
& Third Ventricle &  0.61$_{\pm\text{0.07}}$ & 0.77$_{\pm\text{0.07}}$ \\
Thalamus &  0.73$_{\pm\text{0.05}}$ & 0.86$_{\pm\text{0.04}}$
& Fourth Ventricle &  0.60$_{\pm\text{0.20}}$ & 0.75$_{\pm\text{0.07}}$ \\
Caudate &  0.54$_{\pm\text{0.07}}$ & 0.82$_{\pm\text{0.04}}$
& Brainstem &  0.85$_{\pm\text{0.09}}$ & 0.92$_{\pm\text{0.02}}$ \\
Putamen       &  0.72$_{\pm\text{0.05}}$ & 0.84$_{\pm\text{0.03}}$
& Accumbens &  0.42$_{\pm\text{0.11}}$ & 0.66$_{\pm\text{0.10}}$ \\
Lat. Ventricle &  0.66$_{\pm\text{0.10}}$ & 0.82$_{\pm\text{0.05}}$
& Ventral DC &  0.71$_{\pm\text{0.07}}$ & 0.83$_{\pm\text{0.07}}$ \\
\hline\hline
\end{tabular}
\label{tab:region_dice}
}
\end{table}

\begin{figure}[t]
\centerline{\includegraphics[width=\columnwidth]{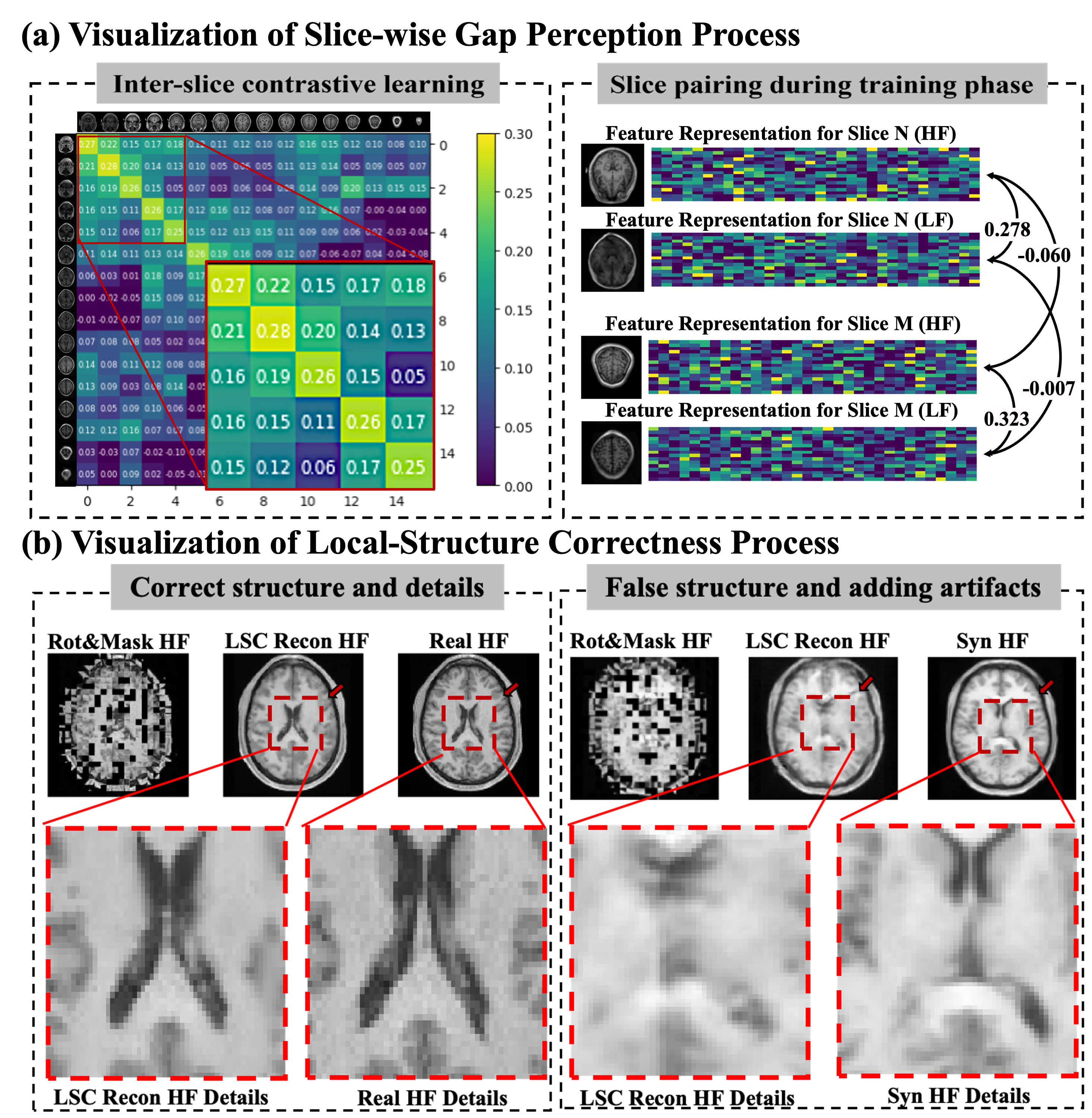}}
\caption{Visualization result of SGP and LSC process. 
(a) SGP enhances inter-slice similarity by pre-training on sequential low-field and high-field MRI data and matching the most similar slices within a randomly shuffled batch. (b) LSC enhances local structures by recovering fine image details from locally masked and rotated images.}
\label{fig3}
\end{figure}

\begin{figure}[t]
\centerline{\includegraphics[width=\columnwidth]{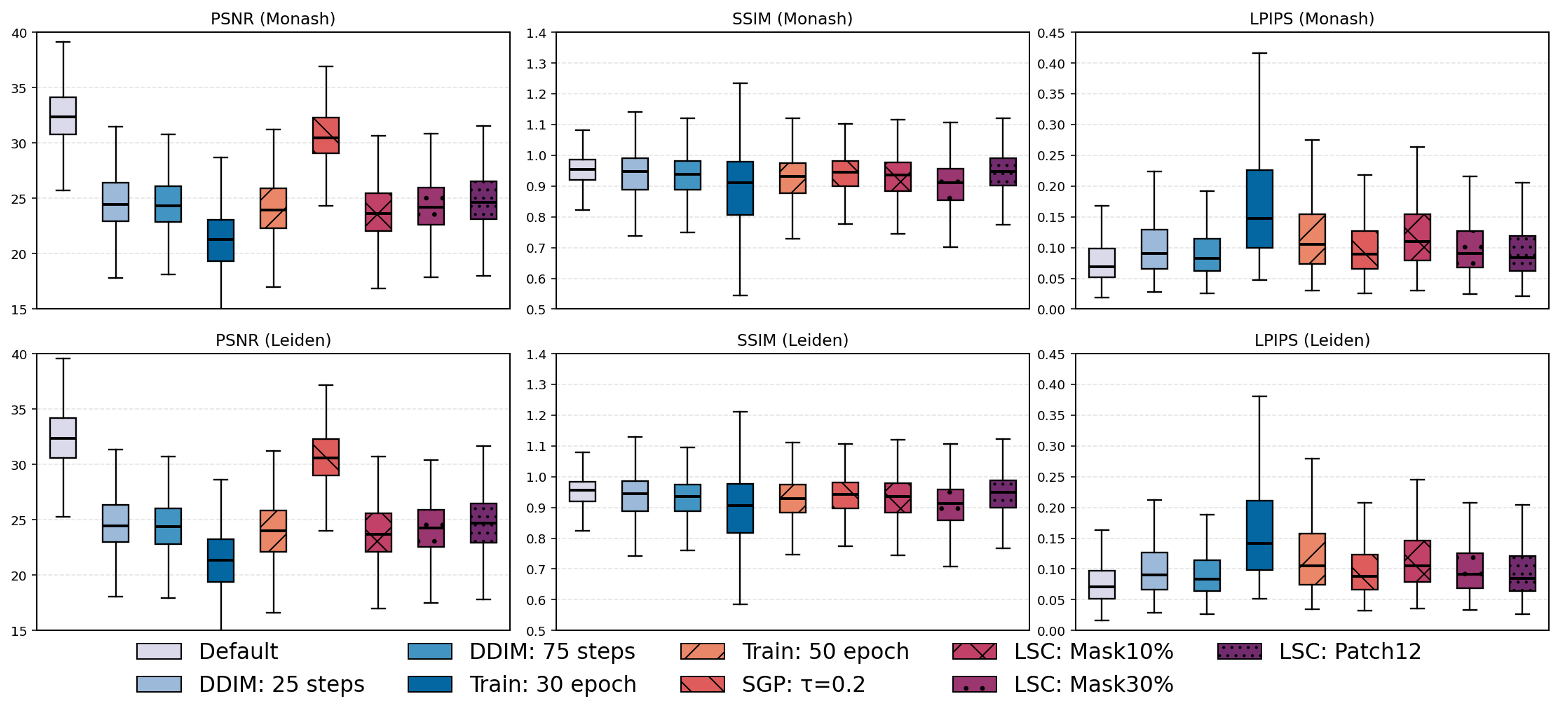}}
\caption{
Ablation study on sampling and network parameters of the CSS-Diff framework, evaluated on the paired 64 mT $\rightarrow$ 3 T dataset.}
\label{fig9}
\end{figure}

\begin{figure}[t]
\centerline{\includegraphics[width=\columnwidth]{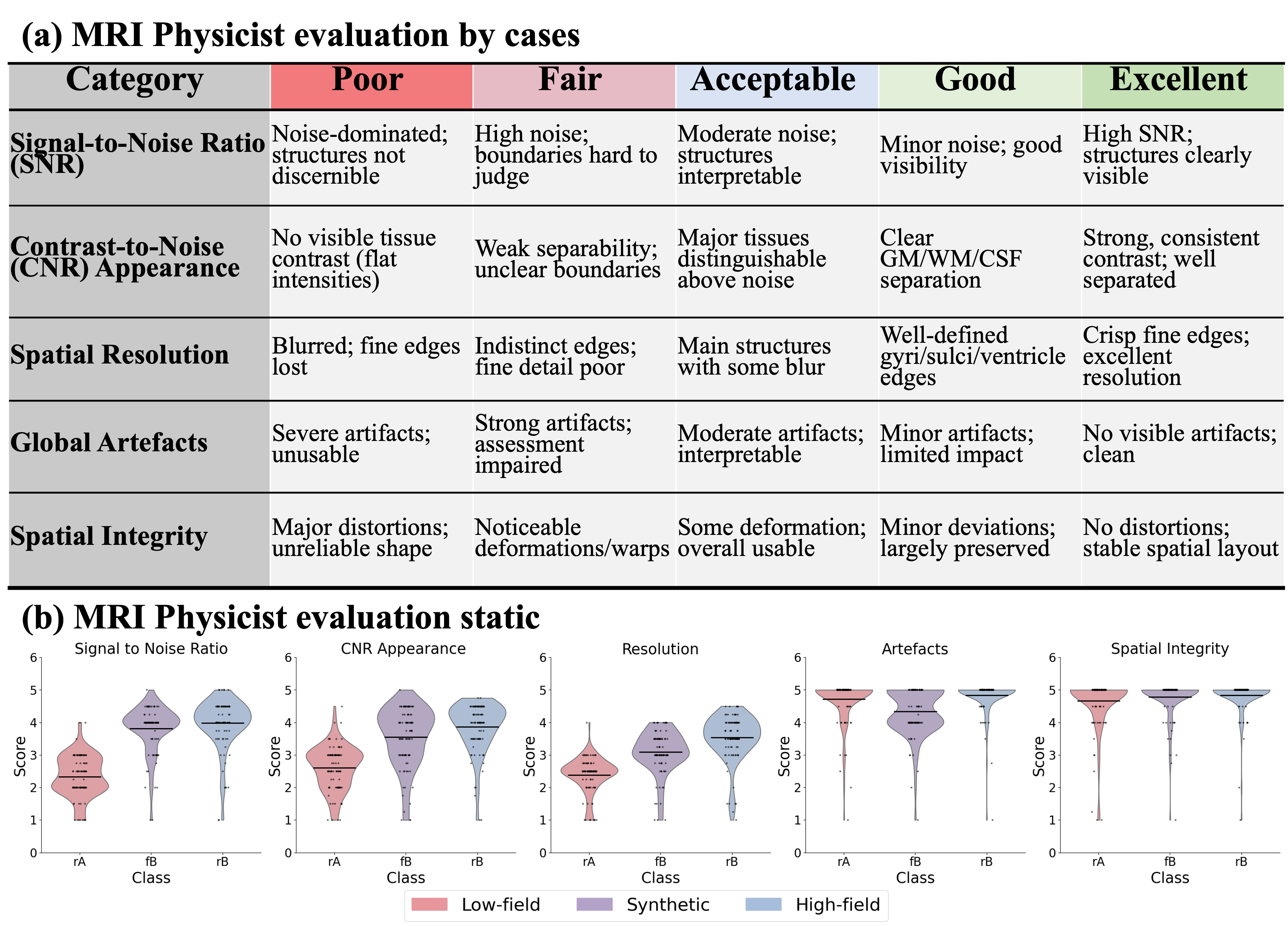}}
\caption{MRI physicist evaluation of our CSS-Diff. (a) MRI physicist evaluation study, comparing quality scores between low-field MRI and synthesized MRI using a 5-point Likert scale. (b) MRI physicist evaluation statistics, including box plots, histograms, and correlation analysis.}
\label{fig6}
\end{figure}

\begin{figure}[t]
\centerline{\includegraphics[width=\columnwidth]{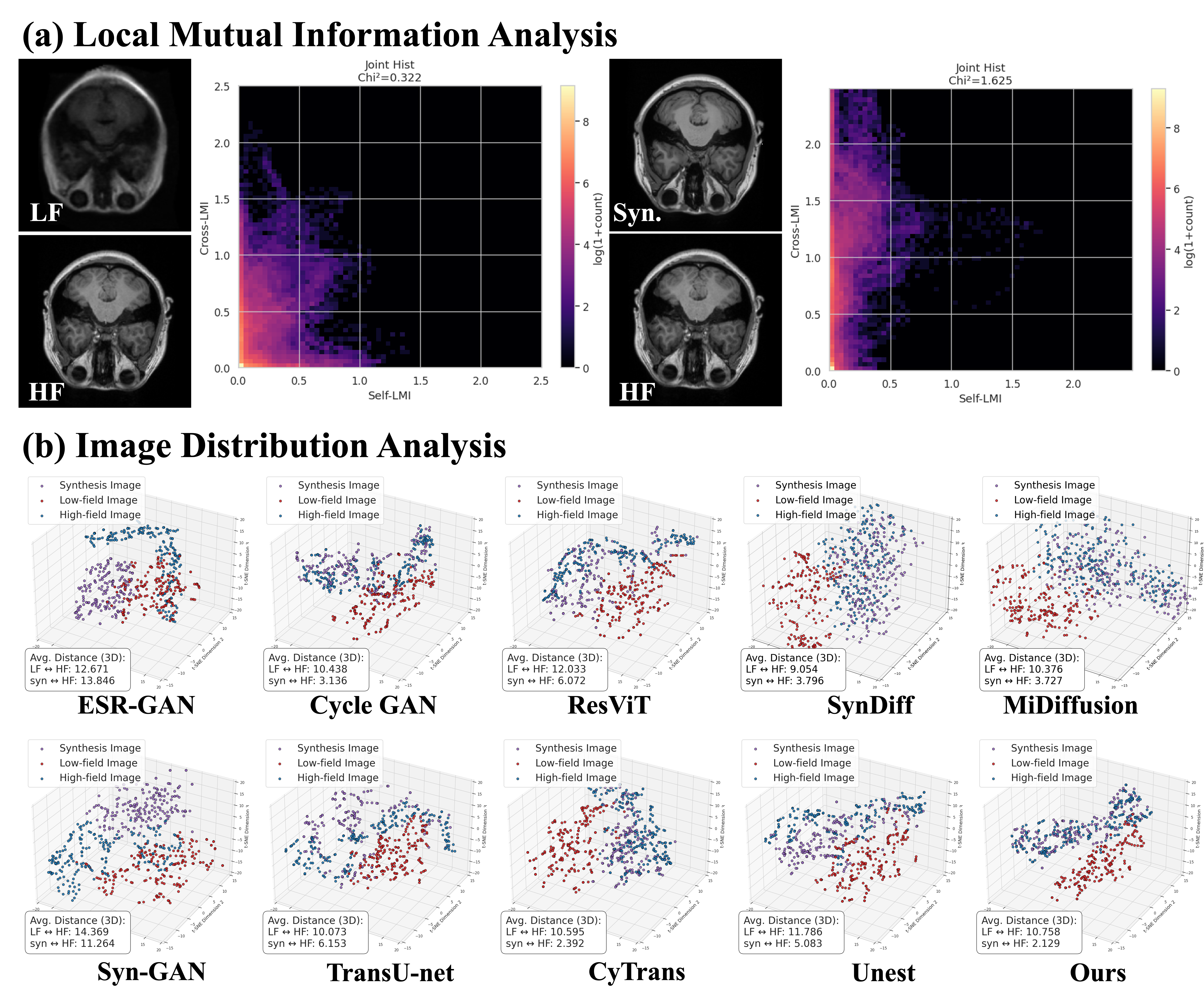}}
\caption{
(a) Joint Local Mutual Information (LMI) distributions showing that our method enhances alignment with real high-field MRI and yields clearer case-wise separability.
(b) \textit{t}-SNE visualization of synthesized (purple) and real high-field (blue) MRI slice features across different methods. Shaded contours indicate feature density, and statistics report the average inter-domain distance and overlap.}
\label{fig:unpaired}
\end{figure}

\begin{figure*}[t]
\centerline{\includegraphics[width=2.05\columnwidth]{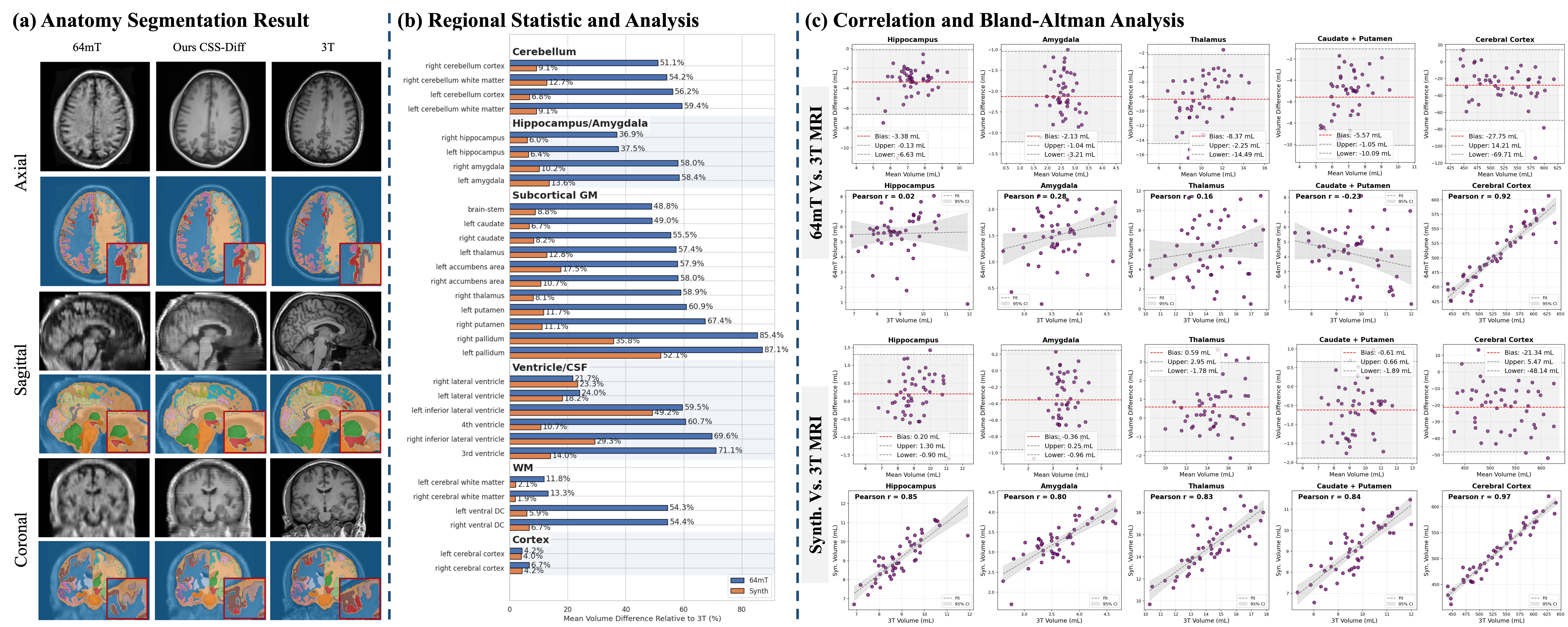}}
\caption{Anatomical segmentation improvement via synthesized high-field MRI. (a) Visual comparison of anatomical structure segmentation using low-field (64mT) MRI, our synthesized images (CSS-Diff), and high-field (3T) MRI. (b) Quantitative analysis of mean volume difference per brain region, comparing segmentations from low-field and synthesized images against the high-field reference. (c) Quantitative agreement between 64mT and 3T brain volume estimations across different structures and contrasts. Bland–Altman plots showing volume differences (64mT$-$3T) for the left hippocampus, left cerebral white matter, and right cerebellum cortex across T$_1$w, T$_2$w, and FLAIR. Correlation analysis plots with Pearson's r values, highlighting the structural consistency or divergence across field strengths.}
\label{fig_seg}
\end{figure*}

\section{Experiment}
\subsection{Dataset and Implementation Description:} 
\subsubsection{Dataset} consists of three sources. The first is a private collection of 20 cases with T$_1$w, T$_2$w, and FLAIR scans acquired at both 64 mT and 3 T. The second is the public Leiden University dataset comprising 11 healthy subjects scanned at both 64 mT and 3 T, including localizer, T$_1$w, T$_2$w, FLAIR sequences, with high-field acquisitions at both standard clinical resolution and resolution matched to the low-field scans \cite{leiden_MRI}. The third is an external dataset from King’s College London (KCL), including 23 healthy participants scanned with both 3 T and 64 mT systems \cite{KCL_dataset}. For the 64 mT acquisitions, two protocols were used: ses-HFC, acquired at the Centre for Neuroimaging Sciences on the same day as the 3 T scans, and ses-HFE, acquired at the Evelina Newborn Imaging Centre within 36 days. Both protocols included T$_1$w and T$_2$w scans.

For all paired datasets, preprocessing includes rigid registration of low-field scans to their high-field counterparts on a per-subject basis. Before registration, images are resampled to isotropic 1~mm resolution to ensure consistent voxel geometry. We use the 3T T$_2$w image as the fixed reference for alignment and apply rigid-body transformation to all other modalities. This process corrects for head motion, scanner-specific geometry distortions, and inter-slice spacing differences, enabling accurate spatial correspondence across field strengths and contrasts.

\subsubsection{Evaluation Metrics} To evaluate CSS-Diff, we use three different metrics. We use the Peak Signal-to-Noise Ratio (PSNR) and Structural Similarity Index Measure (SSIM) to assess pixel-level accuracy and perceptual similarity, while Learned Perceptual Image Patch Similarity (LPIPS) measures deep feature–space differences. 

\subsubsection{Experimental Settings} Our experimental setup employs the Adam optimizer to minimize the joint loss of CSS-Diff, starting with a learning rate of 0.002, which is halved every 10 epochs. Training proceeds for up to 120 epochs, with early stopping (patience = 5) based on validation PSNR. A dropout rate of 0.2 is used to mitigate overfitting. All experiments were conducted on a workstation equipped with a 2.90~GHz Xeon CPU and an NVIDIA H100 GPU.

\subsection{Comparison Experiment} 
Table~\ref{tab:comparison} compares the proposed CSS-Diff framework with a series of paired and unpaired MRI synthesis methods on the paired 64\,mT$\rightarrow$3\,T dataset under the multi-contrast (T$_1$w, T$_2$w, FLAIR) setting. This benchmarked against 3 representative unpaired methods (SynGAN \cite{SynGAN}, CycleGAN \cite{Cyclegan}, UNest \cite{UNest}) and 5 paired methods (Pix2Pix \cite{isola2017image}, ESRGAN \cite{esrgan}, ResViT \cite{resvit}, TranUnet \cite{transunet}, CyTran \cite{cytran}), covering both adversarial and supervised paradigms. Recent diffusion-based models (SynDiff \cite{SynDiff}, MiDiffusion \cite{MiDiffusion}) were also included to reflect the latest generative advances. Across the internal Monash (private) and Leiden (public) datasets, CSS-Diff achieves the best PSNR/SSIM and the lowest LPIPS, indicating superior fidelity and perceptual quality. On Monash, it attains 31.80 dB PSNR, 0.943 SSIM, and 0.0864 LPIPS; on Leiden, it reaches 31.96 dB PSNR, 0.948 SSIM, and 0.0850 LPIPS.
External testing on the KCL ses-HFE and ses-HFC datasets further shows reasonable generalization under distribution shift: CSS-Diff achieves 27.25 dB / 0.807 / 0.1901 and 26.97 dB / 0.785 / 0.2006 (PSNR/SSIM/LPIPS), respectively, outperforming both paired and unpaired baselines. These gains are consistent across settings, demonstrating the robustness of our CSS-Diff. Fig.~\ref{fig4} shows that CSS-Diff produces images with sharper textures and more faithful anatomical structures than other methods, as indicated by red arrows. The model effectively reduces structural distortions and preserves fine details, confirming its advantage in high-field MRI synthesis from low-field inputs.

\subsection{Ablation Study} 
\subsubsection{Effectiveness Evaluation of Different Modules} 
Fig.~\ref{fig3} visualizes the role of SGP and LSC module.  Fig.~\ref{fig3} (a) shows that by pairing slices during training, the model learns to embed sequence-level information that can be injected into the generation phase. The consistency of these feature representations indicates that the network has indeed captured inter-slice dependencies. Fig.~\ref{fig3} (b) highlights that LSC explicitly magnifies and corrects unrealistic or blurred structures in the synthesis process. This ensures that the generated images not only look realistic but also maintain anatomical reliability. Table~\ref{tab:module_ablation} further numerically shows the effectiveness of the SGP, LSC and CCD modules. For the Monash dataset, the full configuration achieves the highest PSNR (32.44), best SSIM (0.940), and lowest LPIPS (0.0857). For the Leiden dataset, it likewise delivers the top performance with PSNR of 32.38, SSIM of 0.937, and LPIPS of 0.0829. For the KCL dataset, the complete CSS-Diff improves PSNR to 27.25 dB and SSIM to 0.807. These results highlight its generalizability under distribution shift across different data sources and confirm that SGP, LSC, and CCD are jointly ensuring superior synthesis quality.

\subsubsection{Effectiveness Evaluation of Network Configuration} 
Fig.~\ref{fig9} compares PSNR, SSIM, and LPIPS for the under different network configurations. The default setting achieves the best overall balance with high PSNR (26.42 compared to 26.36) and SSIM (0.940 compared to 0.937). DDIM-75 yields slightly lower PSNR (24.42 compared to 24.41) but competitive LPIPS (0.0984 compared to 0.0965), indicating a trade-off between fidelity and perceptual quality. Among LSC variants, Mask30$\%$ reduces LPIPS but at the cost of PSNR, while Patch12 preserves reasonable PSNR with modest SSIM gains.

\section{Discussion}
\subsubsection{Physicist Evaluation Reveals Perceptual Improvements and Limitations}
Fig.~\ref{fig6} (a) shows expert ratings of image quality across five criteria (i.e., signal-to-noise ratio (SNR), contrast-to-noise ratio (CNR), spatial resolution, global artifacts, and spatial integrity) using a 5-point Likert scale from Poor to Excellent. Higher scores indicate clearer tissue contrast, sharper anatomical detail, fewer artifacts, and better spatial fidelity. Fig.~\ref{fig6} (b) shows that synthetic images were rated significantly higher than low-field in SNR (3.82 vs 2.34), CNR (3.55 vs 2.61), and spatial integrity (4.78 vs 4.66) by Wilcoxon signed-rank tests ($p<0.05$), while remaining statistically comparable to high-field on these metrics ($p>0.05$). In addition, the limited effectiveness on the artifact score (4.33 vs 4.72 for synthetic vs low-field) likely stems from subtle speckle-like noise occasionally introduced during synthesis, which physicists perceived as residual artifacts. This could be addressed by incorporating explicit noise modeling or local regularization to suppress speckle. These results further indicate that synthetic images are not only quantitatively superior but also perceived as higher quality by humans, making them closer to real-world applicability in clinical practice.

\subsubsection{CSS-Diff Improves Structural Detail and Alignment with High-field MRI}
Fig.~\ref{fig:unpaired} (a) shows local mutual information analysis between low-field, high-field, and synthesis images. Compared with low-field inputs, the synthesis images exhibit lower intra-class mutual information, indicating more distinguishable internal structures. The cross mutual information with high-field data increases, demonstrating stronger alignment with the target domain. This is visible not only from the numerical metrics ($\chi^2$ increases from 0.322 to 1.625) but also from the histogram distributions, where synthesis images show more concentrated and better aligned patterns with high-field references. Fig.~\ref{fig:unpaired} (b) shows t-SNE plots. Our method produces compact and well-separated clusters similar to real data. Other methods show overlap or distorted shapes. The tighter embeddings achieved by our CSS-Diff demonstrate its ability to maintain class integrity while enhancing cross-domain consistency. These results indicate that CSS-Diff enhances structural discriminability within low-field images while simultaneously improving alignment with high-field distributions. This dual effect in both local statistics and global embeddings, suggests that our method achieves finer intra-class detail preservation and cross-domain consistency.

\subsubsection{Synthetic MRI Improves Anatomical Fidelity for Downstream Clinical Analysis}
Fig.~\ref{fig_seg} (a) shows segmentation and volumetric analysis across modalities. Compared to 64mT, the synthesized images yield clearer structural delineation, especially in the hippocampus, thalamus, and cortex.  Fig.~\ref{fig_seg}(b) shows mean volume differences across brain regions relative to 3T MRI. The largest improvements with synthesis are observed in the cerebellum, hippocampus, amygdala, and subcortical gray matter, where errors drop from over 50$\%$ at 64mT to below 15$\%$. Ventricular volumes, which were substantially overestimated in low-field scans, are also corrected to a large extent. White matter and cortical volumes show smaller initial discrepancies, yet synthesis further reduces these errors. Overall, the results suggest that the proposed method not only corrects systematic biases in deep and small structures but also improves consistency in large-scale anatomy. Fig.~\ref{fig_seg} (c) further validates this via Bland–Altman plots and Pearson correlations, showing reduced bias and tighter confidence intervals (e.g., cortex from 0.92 to 0.97, hippocampus from 0.02 to 0.85). Dice scores in Table~\ref{tab:region_dice} also improve consistently, with cerebral GM from 0.65 to 0.75, caudate from 0.54 to 0.82, and thalamus from 0.73 to 0.86. These results suggest a low rate of synthesis hallucination. Nonetheless, gains are more modest in cortex and white matter, especially near periventricular boundaries and highly folded cortical ribbon where subtle sulci and small vessels remain challenging.

\section{Conclusion}


In this work, we propose a cyclic self-supervised diffusion (CSS-Diff) framework that transforms low-field inputs into high-field MRI, incorporating slice-wise gap perception and local structure correction to enhance anatomical fidelity. On low-to-ultra-high-field synthesis, CSS-Diff achieved superior performance compared with baseline methods. These results highlight its potential for generating high-quality, high-field–like MRI data to augment downstream tasks and extend to other imaging modalities.

%
%
%
\bibliographystyle{ieeetr}
\bibliography{Journal_version/ref}
\end{document}